\documentclass[10pt,twocolumn,letterpaper]{article}

\usepackage[letterpaper,top=0.9in,bottom=1in,left=0.75in,right=0.75in,columnsep=0.25in]{geometry}
\usepackage[T1]{fontenc}
\usepackage{times}
\usepackage[dvipsnames]{xcolor}
\usepackage{graphicx}
\usepackage[hyphens]{url}
\usepackage[numbers,sort&compress]{natbib}
\usepackage[font=small,labelfont=bf]{caption}
\usepackage{subcaption}
\usepackage[shortlabels,inline]{enumitem}
\usepackage{etoolbox}

\newcommand{\supp}[1]{\cref{#1}}
\newcommand{\thesupp}[1]{\cref{#1}}
\usepackage{booktabs}
\usepackage{colortbl}
\usepackage{multirow}
\usepackage{amsmath,amssymb}
\usepackage{xspace}
\usepackage{tikz}
\usetikzlibrary{positioning,arrows.meta,fit,calc,backgrounds}

\newcommand{\ours}{PixelTriage\xspace}
\newcommand{\sone}{System One\xspace}
\newcommand{\stwo}{System Two\xspace}
\newcommand{\pvis}{p_{\mathrm{vis}}}
\newcommand{\mthree}{M$^3$Exam\xspace}
\newcommand{\dmv}{DMV\xspace}
\newcommand{\memeye}{MemEye\xspace}
\newcommand{\topk}[1]{PixelTriage@#1}
\definecolor{sonecolor}{HTML}{C0392B}
\definecolor{rankcolor}{HTML}{2C7FB8}
\definecolor{lightgray}{gray}{0.93}
\definecolor{inkcolor}{HTML}{222222}
\definecolor{midgray}{HTML}{6B6B6B}
\definecolor{panelfill}{HTML}{F5F5F3}
\definecolor{lowcolor}{HTML}{7B9E3E}
\definecolor{okgreen}{HTML}{2E7D32}
\definecolor{badred}{HTML}{B03A2E}
\usepackage{pifont}
\newsavebox{\teaserleft}
\newif\ifarxiv
\providecommand{\supp}[1]{supplementary material}
\providecommand{\thesupp}[1]{the supplementary material}

\arxivtrue

\makeatletter
\renewcommand\section{\@startsection{section}{1}{\z@}{-2.2ex \@plus -0.6ex \@minus -0.2ex}{1.2ex \@plus 0.2ex}{\large\bfseries}}
\renewcommand\subsection{\@startsection{subsection}{2}{\z@}{-1.8ex \@plus -0.5ex \@minus -0.2ex}{0.8ex \@plus 0.2ex}{\normalsize\bfseries}}
\renewcommand\paragraph{\@startsection{paragraph}{4}{\z@}{1.2ex \@plus 0.4ex \@minus 0.2ex}{-0.6em}{\normalsize\bfseries}}
\makeatother
\newcommand{\papertitle}{Decide Before You Look: Learning Which Retrieved Memories Deserve Pixels}
\newcommand{\paperauthors}{Youxing LI\\
  The Chinese University of Hong Kong\\
  {\small\texttt{rickliyouxing1103@gmail.com}}}

\definecolor{linkcolor}{HTML}{1F4E9A}
\usepackage[breaklinks,colorlinks,allcolors=linkcolor]{hyperref}
\hypersetup{pdftitle={\papertitle},pdfauthor={Youxing LI}}
\usepackage[capitalize]{cleveref}

\title{\papertitle}
\author{\paperauthors}
\date{}

\begin{document}
\twocolumn[{%
\renewcommand\twocolumn[1][]{#1}%
\maketitle
\vspace{-1.5em}
\begin{center}
\sbox{\teaserleft}{\begin{tikzpicture}[x=1cm, y=1cm, font=\scriptsize,
  card/.style={rounded corners=2pt, line width=0.8pt, fill=white},
  chip/.style={font=\scriptsize\bfseries, inner sep=1.2pt, rounded corners=1pt, text=white}]
  \node[anchor=north west, fill=panelfill, rounded corners=2pt, inner sep=3pt, text width=9.25cm, align=left] (q) at (0,0)
    {\textbf{Q:} Earlier you saw a product that had a \textbf{white orchid stem} on it. Which product was it?\\[-1pt]\mbox{}\hfill\textcolor{midgray}{answer: Vesper Vase 05}};
  \draw[card, draw=rankcolor] (0,-0.93) rectangle (4.62,-2.6);
  \draw[card, draw=sonecolor] (4.82,-0.93) rectangle (9.44,-2.6);
  \node[anchor=north west, inner sep=0] at (0.1,-1.03) {\includegraphics[height=1.47cm]{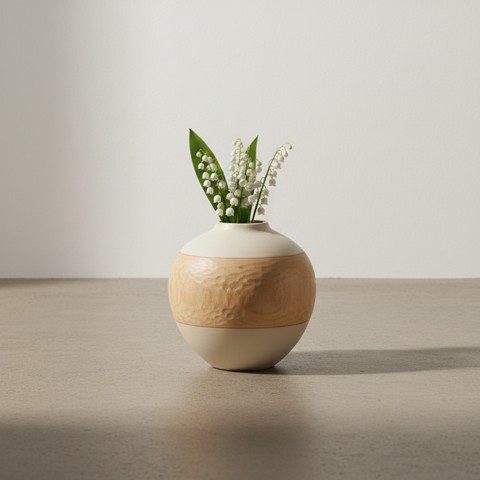}};
  \node[anchor=north west, inner sep=0] at (4.92,-1.03) {\includegraphics[height=1.47cm]{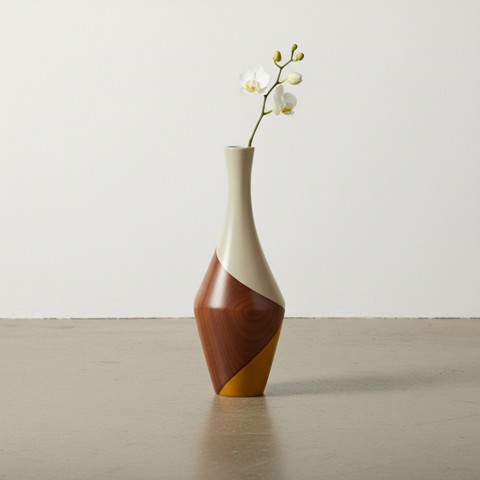}};
  \node[anchor=north west, text width=2.85cm, align=left, inner sep=0] at (1.68,-1.05)
    {\textbf{\#1} \ Aria Vase 05\\[1.5pt]\textcolor{midgray}{\textit{proxy:} ``a spherical vase with a two-tone design \dots\ a band of light brown wood''}};
  \node[anchor=north west, text width=2.85cm, align=left, inner sep=0] at (6.5,-1.05)
    {\textbf{\#3} \ Vesper Vase 05\\[1.5pt]\textcolor{midgray}{\textit{proxy:} ``a tall, slender vase \dots\ beige, dark brown wood grain, mustard yellow''}};
  \node[chip, fill=rankcolor, anchor=south west] at (0.1,-0.92) {retrieval order opens \#1};
  \node[chip, fill=sonecolor, anchor=south west] at (4.92,-0.92) {PixelTriage opens \#3};
  \def\bx{3.05}\def\bw{2.7}
  \node[anchor=west, text=midgray] at (\bx,-2.85) {visual tokens};
  \node[anchor=west, text=midgray] at (6.55,-2.85) {answer};
  \foreach \y/\name/\pct/\col/\ans/\ok in {
      -3.17/text proxies only/0/midgray/``None''/0,
      -3.5/retrieval order (opens \#1)/10/rankcolor/``Hollow Pine Vase 05''/0,
      -3.83/all ten images down-sized/10/lowcolor/``Hollow Pine Vase 05''/0,
      -4.16/\textbf{PixelTriage (opens \#3)}/10/sonecolor/``Vesper Vase 05''/1,
      -4.49/open all ten images/100/midgray/``Vesper Vase 05''/1} {
    \node[anchor=west, inner sep=0] at (0.05,\y) {\name};
    \fill[gray!14] (\bx,\y-0.085) rectangle ++(\bw,0.17);
    \ifnum\pct>0 \fill[\col] (\bx,\y-0.085) rectangle ++({\bw*\pct/100},0.17); \fi
    \node[anchor=west, inner sep=0, text=midgray] at ({\bx+\bw+0.08},\y) {\pct\%};
    \ifnum\ok=1
      \node[anchor=west, inner sep=0] at (6.55,\y) {\ans~\textcolor{okgreen}{\ding{51}}};
    \else
      \node[anchor=west, inner sep=0] at (6.55,\y) {\ans~\textcolor{badred}{\ding{55}}};
    \fi
  }
\end{tikzpicture}}%
\begin{minipage}[b]{0.555\textwidth}
\centering
\usebox{\teaserleft}
\end{minipage}\hfill
\begin{minipage}[b]{0.425\textwidth}
\centering
\includegraphics[height=\dimexpr\ht\teaserleft+\dp\teaserleft\relax]{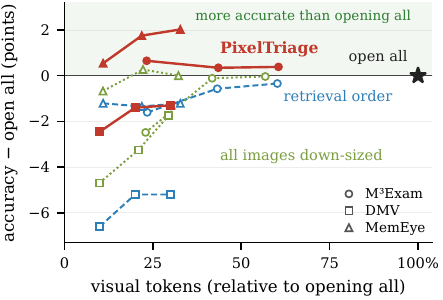}
\end{minipage}
\captionof{figure}{Left: the answer is visible only in the photo, and \ours opens the right one at 10\% of the visual tokens. Right: accuracy relative to opening all retrieved images at budgets $k{=}1,2,3$ (7B answering model).}
\label{fig:teaser}

\vspace{0.8em}
\end{center}
}]
\begin{abstract}
Multimodal assistants answer questions from long-term memories that contain images.
After retrieval, each retrieved image reaches the answering model either as pixels, at about a thousand visual tokens per image, or as a stored text proxy that often misses the detail the question asks about.
We find that the benefit of pixels usually comes from one or two retrieved memories, and that it can be predicted before the answering model runs, without reading any full-resolution image.
In PixelTriage, a plug-in placed after retrieval, a small model that does not generate text reads the dialogue, a short note and a thumbnail of each retrieved memory and predicts how much its pixels would add.
It is trained on synthetic memory episodes labeled by a frozen 27B model that answers each question with and without each memory's pixels.
With a 7B answering model, PixelTriage lies on the accuracy--cost frontier of M$^3$Exam, DMV and MemEye and uses 11--23\% of the visual tokens without a significant loss of accuracy. On DMV it answers 2.9 times faster than opening all images.
It outperforms retrieval order and uniform down-sizing at equal budgets and transfers to other memory systems and to a 397B answering model.
\end{abstract}

\section{Introduction}
\label{sec:intro}

Assistants built on multimodal large language models keep long-term memories of what their users share, such as photos, screenshots, product pages and documents~\cite{long2025seeing,bei2026memgallery,guo2026memeye,huang2026m3exam}.
A memory system stores each image as pixels and as a short text proxy, for example a caption.
After retrieving a few memories for a question, it must decide for each retrieved image whether the answering model sees the pixels or only the text.

Both defaults are costly.
Ten retrieved images take about 12k visual tokens on \dmv-Bench, which slows every answer, and the cost grows with retrieval depth.
Text proxies drop details that nobody wrote down, like the orchid stem in \cref{fig:teaser}.
On \dmv, answering from proxies alone loses 10.3 accuracy points with a 7B answering model and 22.4 points with a 397B model, and recent work finds that delivery can matter more than retrieval~\cite{jiang2026retrieved}.

Existing systems control this choice either with fixed rules or with a generative model.
Rules are cheap but choose for the wrong reason.
Retrieval order measures how related a memory is to the question, which is not what its proxy is missing, and down-sizing every image blurs the detail in question.
Asking the answering model to pick images is flexible but adds a generative call to every question before answering begins.
Methods that inspect the evidence first, by scoring full-resolution images with a surrogate model~\cite{luo2026utility} or by escalating to images after a text-only answer~\cite{huang2026m3exam}, work at a later stage and can run after an earlier, cheaper decision.

We view delivery through the dual-process distinction between fast, automatic decisions and slow, deliberate reasoning~\cite{kahneman2011thinking}.
Choosing which retrieved memories to open is a structured, high-frequency decision with a bounded output, a probability per memory.
It needs semantic understanding of the question and the memories but no text generation, so it fits a fast \sone.
Composing the answer from the delivered evidence remains the work of a \stwo, the answering model, which then runs once.
Decision models that output typed probabilities instead of text~\cite{valen} make such a \sone practical, and two observations tell us what it should predict and how to supervise it without annotation.

First, the benefit of pixels is concentrated: in our development labels, 87\% of the questions where pixels help have at most two of 20 candidates with more than half of the largest gain, and on three benchmarks one image picked by our model recovers most of the gain of opening all retrieved images (\cref{fig:recovery}).
\sone therefore only needs to find one or two memories.
Second, the gain of a memory's pixels can be measured directly, by swapping its proxy for its pixels in front of a frozen answering model, so training labels need no annotation.

\ours (\cref{fig:method}) implements this as a plug-in between retrieval and answering.
Its \sone, a 2B vision-language model, reads the dialogue text, a short note and a 336-pixel thumbnail of each retrieved memory and predicts a pixel gain per memory in one forward pass.
We train it on 9.8k synthetic multi-session episodes built from single-image VQA and labeled by a frozen Qwen3.5-27B (\cref{sec:method:data}).
Among 27 delivery policies on \mthree, \dmv and \memeye, \ours lies on the accuracy--cost frontier of every benchmark.
On \mthree and \memeye, one opened image is more accurate than opening all images at 23\% and 11\% of the visual tokens, and on \dmv \ours answers 2.9 times faster than opening all ten images, including its own cost.

Our contributions are as follows.
\begin{itemize}
\item We cast the delivery of retrieved images as a pixel-budget decision made before answering and show that the benefit of pixels concentrates in one or two memories.
\item We introduce \ours, a non-generative \sone that decides delivery without reading full-resolution images, trained with counterfactual labels from a frozen answering model on memory episodes built from single-image VQA.
\item With a 7B answering model, \ours cuts visual tokens to 11--23\% without a significant accuracy loss and outperforms retrieval order, self-selection and down-sizing at equal budgets. It transfers to two other memory systems and to a 397B answering model.
\end{itemize}

\section{Related Work}
\label{sec:related}

\paragraph{Multimodal memory systems and benchmarks.}
Multimodal agents store images with the dialogue and retrieve them later.
MuRAG~\cite{chen2022murag} retrieves image--text memories for generation, UniversalRAG~\cite{yeo2025universalrag} routes retrieval across corpora of different modalities and granularities, V-Mem~\cite{kang2026vmem} routes retrieval by the modality of the target evidence, and M3-Agent~\cite{long2025seeing} builds long-term memory from continuous multimodal input.
Benchmarks such as Mem-Gallery~\cite{bei2026memgallery}, MemEye~\cite{guo2026memeye}, \mthree~\cite{huang2026m3exam}, \dmv-Bench~\cite{tang2026dmvbench} and SMMBench~\cite{smmbench2026} test whether agents recall visual details from long histories.
\ours takes the retrieved set of such a system as given and decides only in which form each retrieved image reaches the answering model.

\paragraph{Delivery and selection of retrieved visual evidence.}
DeliverMem~\cite{jiang2026retrieved} separates delivery from retrieval in multimodal memory and finds that delivering pixels matters more than improving retrieval. It delivers the pixels of every retrieved memory.
Utility-oriented evidence selection~\cite{luo2026utility} argues that the utility of visual evidence differs from its relevance and estimates it with a surrogate multimodal model that reads each candidate image at full resolution, and drops the images it does not select.
M3Proctor~\cite{huang2026m3exam} answers from text surrogates first and escalates to raw images when its confidence is low, deciding per question with additional answering calls.
Submodular selection~\cite{sharma2025submodular} chooses image subsets for multi-image question answering by relevance.
VisRAG~\cite{yu2025visrag} and MRAG-Bench~\cite{hu2025mragbench} show that pixel evidence can be more useful than its textual description.
These methods refine the evidence after reading full-resolution images or after a first answer.
\ours decides earlier, per memory and before the answering model runs, from a thumbnail and a short note of each memory instead of the full-resolution image. It keeps the text proxy of every unopened memory and learns its decision from counterfactual responses.
The two stages are complementary, and \ours can run first to limit which images the later methods read.

\paragraph{Visual token reduction.}
FastV~\cite{chen2024fastv} and SparseVLM~\cite{zhang2025sparsevlm} prune visual tokens inside the model, RUTA~\cite{zou2026ruta} allocates token budgets across image--query pairs with a rate--utility objective, and consequence-sensitive compression~\cite{wen2026consequence} adapts the compression rate of each image to the question.
They shrink every delivered image, whereas \ours chooses which retrieved memories receive pixels and leaves the rest as text. The two can be combined, and the matched-budget down-sizing baseline in \cref{sec:exp:frontier} compares the two allocations at equal token counts.

\paragraph{Adaptive retrieval and decision models.}
Self-RAG~\cite{asai2024selfrag} and Adaptive-RAG~\cite{jeong2024adaptiverag} decide whether and how to retrieve for each query.
Non-generative decision models that separate typed decisions from slow generation have been used to control the construction and retrieval of text memories~\cite{jiang2026jevmem}, and Valen~\cite{valen} extends such decision models to visual input and provides our backbone and initialization.
We apply this separation at a different step, after retrieval and before answering, and train the decision on counterfactual responses of a frozen answering model, which general decision data does not provide (\cref{sec:exp:b1}).

\section{Problem Setting}
\label{sec:problem}

\paragraph{Delivery after retrieval.}
A host memory system stores every image memory $m_i$ in two forms: its pixels $x_i$ and a text proxy $t_i$, such as a caption or a note written when the image was saved, together with the dialogue text $u_i$ of the turn in which it was shared.
Given a question $q$, the host retrieves $n$ memories $R(q)=\{m_1,\dots,m_n\}$.
Before the answering model runs, a delivery $d\in\{0,1\}^n$ fixes the form of every retrieved memory: if $d_i=1$, memory $i$ enters the prompt with its pixels, placed next to its text proxy unless the host's prompt format replaces the proxy, and otherwise it enters as its text proxy alone.
The answering model, which we call \stwo, is then called once on $(q,R(q),d)$.
We write $\mathrm{acc}(d)$ for the judged accuracy of its answer and define the visual cost of a delivery as
\begin{equation}
  c(d)=\sum_{i=1}^{n} d_i\,\tau(x_i),
  \label{eq:cost}
\end{equation}
where $\tau(x_i)$ is the number of visual tokens \stwo spends on image $x_i$.
Opening all images ($d=\mathbf{1}$) and opening none ($d=\mathbf{0}$) are the two defaults, and a pixel budget $k$ restricts a delivery to $\|d\|_1\le k$.

\paragraph{Evaluation by the accuracy--cost frontier.}
A delivery policy maps each question to a delivery.
We summarize a policy on a benchmark by its mean visual cost, reported relative to opening all images, and its mean accuracy.
Policy $A$ dominates policy $B$ if $A$ costs no more and is at least as accurate, with one of the two strict. The undominated policies form the accuracy--cost frontier.
The cost counts the visual tokens of the answering call.
Extra model calls that a policy makes to choose its images, and the compute of our decision model, are reported separately (\cref{sec:exp:cost}).

\paragraph{Counterfactual pixel gain.}
The quantity a delivery decision needs is how much one memory's pixels add over its text proxy for the question at hand.
Let $A(d)\in[0,1]$ be the probability that a fixed answering model answers $q$ correctly under delivery $d$, and let $e_i$ open only memory $i$.
The counterfactual pixel gain of memory $i$ is
\begin{equation}
  g_i=A(e_i)-A(\mathbf{0}).
  \label{eq:gain}
\end{equation}
It depends on what the proxy $t_i$ omits and on what $q$ asks, so two memories that are equally relevant to $q$ can have very different gains.
At the question level, pixels are needed when the text-only delivery fails and opening the right memory succeeds.

\paragraph{The decision problem.}
\sone predicts the gains and the need for pixels from the dialogue text, a short note and a thumbnail of each memory, never from the full-resolution image, and a policy turns the predictions into $d$ without calling \stwo.

\section{PixelTriage: A Delivery Plug-in}
\label{sec:method}

\begin{figure}[t]
\centering
\begin{tikzpicture}[x=1cm, y=1cm, font=\footnotesize,
  img/.style={draw=gray!70, fill=gray!8, rounded corners=0.6pt, line width=0.4pt},
  box/.style={draw=gray!70, rounded corners=2pt, align=center, inner sep=2pt, line width=0.5pt},
  ours/.style={box, draw=sonecolor!85!black, fill=sonecolor!10, line width=0.7pt},
  frozen/.style={box, draw=rankcolor!80!black, fill=rankcolor!8},
  arr/.style={-{Latex[length=1.6mm,width=1.3mm]}, line width=0.6pt, gray!60!black},
  tarr/.style={-{Latex[length=1.6mm,width=1.3mm]}, line width=0.7pt, dashed, sonecolor!85!black}]
  \newcommand{\imgglyph}[4]{%
    \draw[img, draw=#4] (#1,#2) rectangle ++(#3,#3);
    \fill[#4!45] (#1+0.06*#3/0.42,#2+0.06*#3/0.42) -- ++(0.13*#3/0.42,0.16*#3/0.42) -- ++(0.08*#3/0.42,-0.08*#3/0.42) -- ++(0.06*#3/0.42,0.05*#3/0.42) -- ++(0.06*#3/0.42,-0.13*#3/0.42) -- cycle;
    \fill[#4!45] (#1+0.30*#3/0.42,#2+0.31*#3/0.42) circle (0.045*#3/0.42);}
  \node[box, fill=gray!8, minimum width=2.35cm, minimum height=0.62cm] (host) at (1.15,1.02) {host memory system\\[-1pt]{\scriptsize store, retrieve (unchanged)}};
  \foreach \r/\c in {0/gray, 1/gray, 2/gray, 3/gray} {
    \imgglyph{0.02}{-0.55*\r-0.21}{0.42}{\c}
    \draw[gray!55, line width=1.6pt] (0.58,-0.55*\r+0.10) -- ++(1.62,0);
    \draw[gray!35, line width=1.6pt] (0.58,-0.55*\r-0.07) -- ++(1.15,0);
  }
  \node[font=\scriptsize, gray!40!black, anchor=north, align=center, inner sep=1pt] at (1.12,-1.93) {retrieved memories:\\[-1pt]text, note, thumbnail};
  \draw[arr] (host.south) -- (1.15,0.34);
  \draw[sonecolor!70, dashed, rounded corners=3pt, line width=0.5pt] (2.52,0.52) rectangle (6.12,-2.22);
  \node[font=\scriptsize\bfseries, text=sonecolor!85!black, anchor=south west, inner sep=1pt] at (2.55,0.53) {PixelTriage};
  \node[ours, minimum width=1.32cm, minimum height=2.05cm] (s1) at (3.33,-0.82) {\textbf{System}\\\textbf{One}\\[2pt]{\scriptsize 2B,}\\[-1pt]{\scriptsize one forward}};
  \draw[arr] (2.28,-0.82) -- (s1.west);
  \node[font=\scriptsize, anchor=west, inner sep=0pt] at (4.22,0.30) {$\pvis$};
  \fill[gray!25] (4.72,0.26) rectangle ++(0.80,0.09);
  \fill[sonecolor!80] (4.72,0.26) rectangle ++(0.66,0.09);
  \foreach \r/\len/\col/\tag in {0/0.22/gray!45/text, 1/0.95/sonecolor!80/pixels, 2/0.30/gray!45/text, 3/0.12/gray!45/text} {
    \fill[\col] (4.25,-0.55*\r-0.12) rectangle ++(\len,0.2);
    \node[font=\scriptsize, anchor=west, inner sep=0pt, text=gray!30!black] at (5.27,-0.55*\r-0.02) {\tag};
  }
  \node[font=\scriptsize, anchor=north] at (4.75,-1.86) {$\hat g_i$, top-$k$};
  \draw[arr] (s1.east) -- (4.2,-0.82);
  \node[box, fill=gray!8, minimum width=1.5cm, minimum height=1.25cm] (s2) at (7.13,-0.82) {};
  \node[font=\footnotesize, align=center, anchor=north] at (7.13,-0.24) {\textbf{System Two}};
  \imgglyph{6.55}{-1.20}{0.42}{sonecolor}
  \draw[gray!55, line width=1.4pt] (7.07,-0.92) -- ++(0.70,0);
  \draw[gray!55, line width=1.4pt] (7.07,-1.07) -- ++(0.70,0);
  \draw[gray!55, line width=1.4pt] (7.07,-1.22) -- ++(0.50,0);
  \node[font=\scriptsize, anchor=north] at (7.13,-1.47) {answers once};
  \draw[arr] (6.12,-0.82) -- (s2.west);
\end{tikzpicture}
\caption{\textbf{\ours.} \sone scores the retrieved memories in one forward pass, the top-$k$ are delivered as pixels, and \stwo answers once. Offline, a frozen \stwo provides the training labels.}
\label{fig:method}
\end{figure}
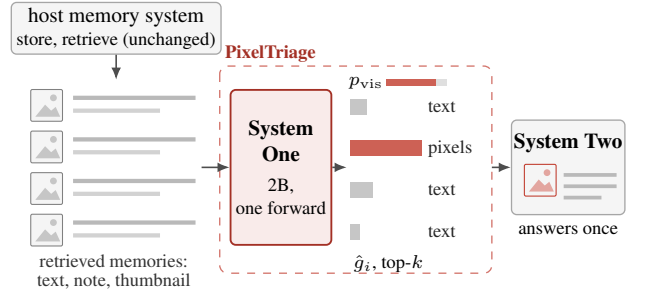

\ours sits between the retrieval of a host memory system and its answering model (\cref{fig:method}).
For every question it decides which retrieved memories reach the answering model as pixels and which as text, using a \sone that predicts the pixel gain of each memory from cheap views of it.
\Cref{sec:method:data} describes how \sone is trained.

\subsection{Interface with the host}
\label{sec:plugin:interface}

From the host, \ours takes the question, the retrieved memories and, for every retrieved image, its pixels, its stored text proxy and the dialogue turn in which it was shared.
It returns a delivery $d$ (\cref{sec:problem}).
The host's storage, index, retrieval, prompt template and answering model stay unchanged.

The only addition is a write-time cache.
When the host stores an image, \ours keeps a 336-pixel thumbnail and a short note keyed by the image hash.
The note is written by a frozen Qwen3.5-27B~\cite{qwen35} without seeing any question and keeps a short label and the first sentence of a caption, about 15 words.
The cache is built once per image, at 2.4 to 3.4\,s per image in our measurements (\supp{sec:supp:write}), and a host that already stores such notes only adds the thumbnails.
At answering time \sone reads this cache and never the full-resolution image.

\subsection{System One: one forward pass over the retrieved set}
\label{sec:method:model}

\sone outputs a question-level probability $\pvis$ that at least one retrieved image must be opened, and a per-memory gain $\hat g_i$ that estimates the counterfactual pixel gain of \cref{eq:gain} on a $[0,1]$ scale.
The delivery policy ranks memories by $\hat g_i$.
A per-memory relevance $\hat r_i$, the probability that memory $i$ is the one the question refers to, is trained as an auxiliary output and is not used at test time.

\paragraph{Backbone and decision head.}
We build \sone on Qwen3.5-2B~\cite{qwen35} with the non-generative decision head of Valen~\cite{valen} in place of the language-model head.
A decision position $h_\mathrm{d}$ is scored against candidate positions $h_j$ with a bilinear form,
\begin{equation}
  \ell_j=\frac{(W_\mathrm{d}h_\mathrm{d})^\top(W_\mathrm{c}h_j)}{\sqrt{D}},
  \label{eq:head}
\end{equation}
with projections of dimension $D=256$.
The candidates of the question-level decision are the answers \emph{true} and \emph{false}.

\paragraph{One state, all memories.}
The input lists the question and then, for each retrieved memory, its session date, the dialogue text in which it was shared, its note and its thumbnail.
Each memory ends with a marker token whose hidden state $h_i$ represents it.
The decision questions follow the memories, so one forward pass yields every output.
Each per-memory question also contains a fixed candidate, \emph{none}, and every memory marker is scored against it,
\begin{equation}
  \hat g_i=\sigma\big(\ell_i-\ell_\mathrm{none}\big).
  \label{eq:item}
\end{equation}
A memory therefore receives a high gain only when it outscores \emph{none}, independently of the other memories, which makes gains comparable across memories and questions.
On the benchmarks a state has 2.2k to 2.4k tokens.

\subsection{Delivery decision}
\label{sec:method:policy}

After retrieval, \sone runs once on the question and the cached views of the retrieved memories.
The policy $\pi_k$ marks the $k$ memories with the highest $\hat g_i$ for pixels and the others for text, and opens every image when fewer than $k$ exist.
The budget $k$ is the operating knob between accuracy and visual cost and needs no retraining.
A gated variant also uses $\pvis$: it opens images only for questions whose $\pvis$ ranks among a chosen fraction of recent questions, which turns $\pvis$ into a question-level budget (\cref{sec:exp:gate}).
Neither variant calls the answering model to decide, and the answering model then runs once on the delivered evidence.
A forward pass of \sone costs about 10\,TFLOP, while the prefill of an answering call that reads all retrieved images costs 97 to 670\,TFLOP in our settings (\cref{sec:exp:cost}).

\subsection{Plugging into different hosts}
\label{sec:plugin:hosts}

\ours applies $d$ in the host's own prompt format.
We consider three cases, all evaluated in \cref{sec:exp}.

\paragraph{Hosts that deliver pixels.}
Multimodal memory systems such as MuRAG~\cite{chen2022murag} hand every retrieved image to the answering model.
\ours keeps the pixels of the selected memories and lets the others fall back to their text proxies, which in MuRAG's prompt are the image identifier and caption.
Whether an opened image keeps its caption or replaces it follows the host's format.
Hosts that route retrieval across modalities, such as UniversalRAG~\cite{yeo2025universalrag}, are handled in the same way on the images they return.

\paragraph{Text-only hosts.}
A text memory system never shows pixels.
\ours leaves its retrieved text and prompt unchanged and appends the pixels of the selected memories, which gives the system visual evidence at the cost of $k$ images.

\paragraph{Other answering models.}
\sone is trained once, on labels from a 27B answering model, and is used unchanged with other answering models, such as the 7B and 397B models of \cref{sec:exp:transfer}.
Only the host's answering call changes.

\section{Training System One Without Annotation}
\label{sec:method:data}

\begin{figure*}[t]
\centering
\begin{tikzpicture}[x=1cm, y=1cm, font=\scriptsize,
  head/.style={font=\footnotesize\bfseries, anchor=north west, inner sep=0pt},
  txt/.style={anchor=north west, inner sep=0pt, align=left},
  arr/.style={-{Latex[length=1.6mm,width=1.3mm]}, line width=0.6pt, gray!60!black},
  th/.style={inner sep=0pt, line width=1.1pt}]
  \node[head] at (0,0) {(1) Source question};
  \node[anchor=north west, inner sep=0pt] (src) at (0,-0.38) {\includegraphics[height=1.55cm]{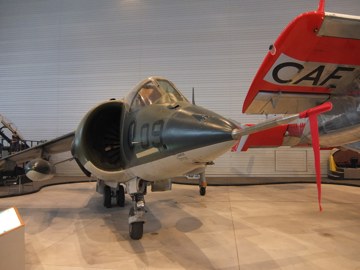}};
  \node[txt, text width=2.45cm] at (0,-2.05) {``What number is written on the nose of this plane?''\\[1pt]\textcolor{gray!40!black}{options: 8, 7, 10, 09}};
  \draw[arr] (2.62,-1.1) -- (2.92,-1.1);
  \node[head] at (3.0,0) {(2) Write-time note};
  \node[txt, text width=2.95cm, draw=gray!60, fill=gray!8, inner sep=2.5pt, rounded corners=1.5pt] (note) at (3.0,-0.42)
    {\textit{text proxy:} military aircraft display. The image shows a military jet aircraft on display in an indoor museum setting.};
  \node[txt, text width=3.0cm, text=gray!40!black, below=4pt of note.south west, anchor=north west] {written by the frozen 27B model without the question (no number), plus a 336-px thumbnail and embeddings};
  \draw[arr] (6.18,-1.1) -- (6.48,-1.1);
  \node[head] at (6.56,0) {(3) Memory episode};
  \def\tw{0.84}\def\thh{0.63}
  \node[th, draw=sonecolor, anchor=north west] at (6.56,-0.4) {\includegraphics[width=\tw cm,height=\thh cm]{figures/assets/pipeline/gold.jpg}};
  \node[th, draw=rankcolor, anchor=north west] at (7.46,-0.4) {\includegraphics[width=\tw cm,height=\thh cm]{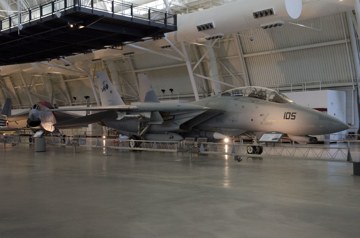}};
  \node[th, draw=rankcolor, anchor=north west] at (8.36,-0.4) {\includegraphics[width=\tw cm,height=\thh cm]{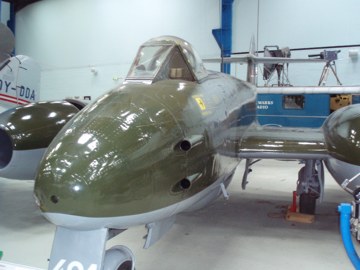}};
  \node[th, draw=rankcolor, anchor=north west] at (9.26,-0.4) {\includegraphics[width=\tw cm,height=\thh cm]{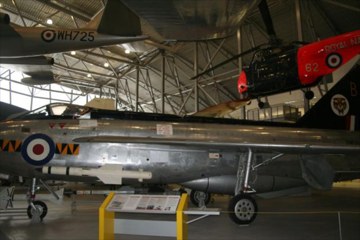}};
  \node[th, draw=rankcolor, anchor=north west] at (6.56,-1.12) {\includegraphics[width=\tw cm,height=\thh cm]{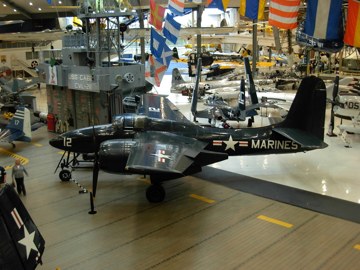}};
  \node[th, draw=violet!70!black, anchor=north west] at (7.46,-1.12) {\includegraphics[width=\tw cm,height=\thh cm]{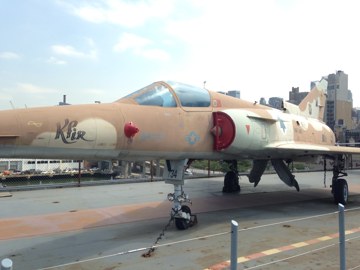}};
  \node[th, draw=violet!70!black, anchor=north west] at (8.36,-1.12) {\includegraphics[width=\tw cm,height=\thh cm]{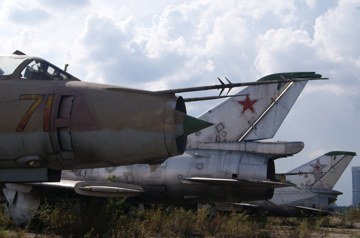}};
  \node[th, draw=gray, anchor=north west] at (9.26,-1.12) {\includegraphics[width=\tw cm,height=\thh cm]{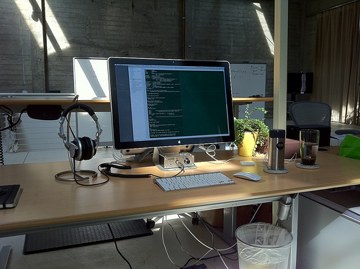}};
  \node[txt, text width=3.6cm] at (6.56,-1.9) {20 memories in 4 dated sessions:\\
    \textcolor{sonecolor}{\textbf{referenced}}, \textcolor{rankcolor}{8 visual neighbors}, \textcolor{violet!70!black}{6 caption neighbors}, \textcolor{gray}{5 random} (8 shown)};
  \draw[arr] (10.22,-1.1) -- (10.52,-1.1);
  \node[head] at (10.6,0) {(4) Recall question};
  \node[txt, text width=3.05cm, draw=gray!60, inner sep=2.5pt, rounded corners=1.5pt] (rq) at (10.6,-0.42)
    {``Earlier I shared the military jet on display next to the red-tailed aircraft. What number is written on the nose of this plane?''\\[2pt]
     A\,24 \quad B\,\textbf{09} \quad C\,05\\D\, There was no such photo};
  \node[txt, text width=3.2cm, text=gray!40!black, below=4pt of rq.south west, anchor=north west] {leakage check, question and options only: $A=0.01$};
  \draw[arr] (13.82,-1.1) -- (14.12,-1.1);
  \node[head] at (14.2,0) {(5) Counterfactual labels};
  \node[txt, text width=3.3cm] at (14.2,-0.38) {frozen 27B model:\\[1pt] $A(T)=0.00$\\ $A(T{+}x_\mathrm{ref})=0.98$\\ $\pvis=0.98$};
  \draw[gray!60, line width=0.4pt] (14.2,-2.45) -- (17.45,-2.45);
  \foreach \k/\g/\c in {1/0.018/violet!70!black, 2/0.037/rankcolor, 3/0.0/gray, 4/0.024/violet!70!black, 5/0.022/rankcolor,
                        6/0.004/rankcolor, 7/0.0/gray, 8/0.001/gray, 9/0.016/rankcolor, 10/0.02/rankcolor,
                        11/0.981/sonecolor, 12/0.011/rankcolor, 13/0.0/gray, 14/0.0/gray, 15/0.009/violet!70!black,
                        16/0.026/violet!70!black, 17/0.022/violet!70!black, 18/0.021/rankcolor, 19/0.007/rankcolor, 20/0.049/violet!70!black} {
    \fill[\c] ({14.22+(\k-1)*0.16},-2.45) rectangle ++(0.11,{max(0.025,\g*0.85)});
  }
  \node[anchor=west, font=\scriptsize, text=sonecolor] at ({14.22+10*0.16+0.13},-1.7) {$g_{\mathrm{ref}}=0.98$};
  \node[anchor=north west, inner sep=0pt, text=gray!40!black, text width=3.15cm, align=left] at (14.2,-2.53) {$g_i$ of the 20 memories\\(others $\le$0.05)};
\end{tikzpicture}
\caption{The training data pipeline on one training query. A frozen 27B model writes the notes and options and labels each memory by the change in the probability of the correct answer when that memory's pixels are added.}
\label{fig:pipeline}
\end{figure*}

\sone predicts counterfactual quantities, the gain of a memory's pixels over its proxy and whether a question needs pixels at all. No annotation of these exists, and memory benchmarks are small and reserved for testing.
We therefore build training data from single-image visual question answering and label it with a frozen answering model in five steps (\cref{fig:pipeline}).
Steps 1 to 4 turn a VQA question into a memory episode (\cref{sec:train:episodes}), step 5 labels every memory (\cref{sec:train:labels}), and \cref{sec:train:objective} trains \sone.

\subsection{Memory episodes from single-image VQA}
\label{sec:train:episodes}

\paragraph{Step 1: source questions.}
We start from the single-image multiple-choice questions of Valen-General-100k~\cite{valen}, taking training questions from GQA~\cite{hudson2019gqa}, VQAv2~\cite{goyal2017vqav2}, TextVQA~\cite{singh2019textvqa} and DocVQA~\cite{mathew2021docvqa}.
ChartQA~\cite{masry2022chartqa} and ScreenQA~\cite{hsiao2025screenqa} are held out as a cross-source development set, and a hashed slice of training-source images forms an in-source development set.
Splits are made at the image level, and every distractor memory comes from the same split as its question.

\paragraph{Step 2: write-time notes.}
As at deployment (\cref{sec:plugin:interface}), the frozen Qwen3.5-27B~\cite{qwen35} writes for every image, without seeing any question, a label of at most five words, a two-to-four-sentence caption and an OCR transcript.
The stored text proxy keeps only the label and the first caption sentence, about 15 words, so that fine detail remains in the pixels.
Each image also gets a 336-pixel thumbnail, a SigLIP~2~\cite{tschannen2025siglip2} image embedding and a Qwen3~\cite{zhang2025qwen3embedding} embedding of its caption, which step 3 uses to find neighbors.

\paragraph{Step 3: memory episodes.}
Each query receives 20 candidate memories spread over four to eight dated sessions.
Besides the image the question is about, there are eight nearest visual neighbors, six nearest caption neighbors and five random images, with near-duplicates (cosine similarity above 0.97) excluded.
Every memory is a dated user turn that shares its image.

\paragraph{Step 4: recall questions and options.}
The original question is rewritten as a recall question that points to the memory by its session date (40\% of queries), by a content phrase written by the 27B model from the label and caption without revealing the answer (40\%), or through another memory shared in the same session (20\%, two hops).
The 27B model also replaces the original distractor options, which often differ in type from the answer, by three options of the answer's type and format.
In 25\% of queries one option states that no such image was shared, an answer that lossy proxies often lead answering models to choose.
As a leakage check, the 27B model answers from the question and options alone: the correct option receives probability 0.37 on average, against 0.25 for chance.
The result is 9,824 training queries and 2,470 development queries (1,474 in-source, 996 cross-source).

\subsection{Counterfactual labels}
\label{sec:train:labels}

In step 5, the frozen 27B model answers each query under controlled contexts.
In the reference context $T$ every candidate appears through its text proxy, and $T{+}x_i$ adds the pixels of candidate $i$, once for each candidate.
As in \cref{eq:gain}, $A(\cdot)$ is the probability of the correct answer, here read as the probability of the correct option letter, renormalized over the four letters, with the assistant turn prefilled by ``Answer:'' so that the next token is a letter.
The soft targets are
\begin{align}
  g_i &= \min\!\big(1,\max(0,\,A(T{+}x_i)-A(T))\big), \\
  \pvis &= \big(1-A(T)\big)\,A(T{+}x_{\mathrm{ref}}), \\
  r_i &= \mathbf{1}[\,i \text{ is the referenced memory}\,],
\end{align}
where $x_{\mathrm{ref}}$ is the image of the referenced memory.
$g_i$ is the expected accuracy gain from opening candidate $i$, and $\pvis$ is the probability that the proxy-only answer is wrong while the answer with the referenced image is right.
Hard labels re-measured under a permuted option order agree in only about 82\% of cases, so we train on soft targets, and development labels average two option orders.
Labeling takes about 26 answering calls per query.
On the training set the referenced image raises $A$ from 0.47 to 0.81 on average, 39.5\% of the queries need pixels under the hard criterion, and a distractor's pixels raise the answer probability by more than 0.1 in 13.9\% of cases and lower it by more than 0.1 in 11.8\%.

\subsection{Objective and training}
\label{sec:train:objective}

Each query becomes one state with the three decision questions.
The per-memory outputs use binary cross-entropy against the soft targets and $\pvis$ uses Valen's decision loss.
\sone starts from a Valen decision model fine-tuned on Valen-General-100k (our reproduction) and is trained for two epochs with LoRA~\cite{hu2022lora} (rank 32) on all linear layers of the language model, together with the vision merger and the decision head.
On the development set, $\pvis$ reaches an AUROC of 0.821 with an expected calibration error of 0.048, and the memory with the highest $\hat g_i$ is the referenced one in 87.8\% of queries.

\section{Experiments}
\label{sec:exp}

\subsection{Setup}
\label{sec:exp:setup}

\paragraph{Plug-in protocol.}
All experiments run inside the Mem-Gallery evaluation harness~\cite{bei2026memgallery}.
The host is MuRAG~\cite{chen2022murag} with GME-Qwen2-VL-2B~\cite{zhang2025gme} retrieval of the top 10 memories, plugged in as described in \cref{sec:plugin:hosts}, and the answering model is Qwen2.5-VL-7B~\cite{bai2025qwen25vl}, called once with the harness prompt.
The benchmark caption of each image is the host's text proxy.
An opened image is placed next to its caption, except on \mthree, whose captions are mostly empty, where we keep the harness default of replacing the caption. Other caption and delivery conditions are in \thesupp{sec:supp:conditions}.
\sone was trained only on the synthetic episodes of \cref{sec:method:data}.

\paragraph{Benchmarks.}
We use benchmarks in which visual detail matters, selected by a criterion fixed before running \ours: opening all retrieved images must improve accuracy over text proxies by at least 5 points.
\mthree~\cite{huang2026m3exam} (654 open-ended questions about two public user personas, 4.2 retrieved images per question), \dmv-Bench~\cite{tang2026dmvbench} (1,000 recall probes over 200 replayed shopping chains, with the queried cue only in a product photo, 10 retrieved images) and \memeye~\cite{guo2026memeye} (371 open-ended questions, 8.8 retrieved images) pass with gains of 7.5, 10.3 and 11.1 points.
\dmv-Bench is interactive in its original form. We replay its browsing chains as static dialogue, so our numbers are not comparable to the original leaderboard.
Mem-Gallery~\cite{bei2026memgallery} (1.7 points) and SMMBench~\cite{smmbench2026} (at most 3.7) do not pass and are analyzed as boundary cases (\cref{sec:exp:boundary}).

\paragraph{Scoring.}
\mthree answers are judged by Qwen2.5-VL-32B with a five-level rubric, and \dmv and \memeye answers by Qwen2.5-72B with the Mem-Gallery rubric (0, 0.5 or 1), which reproduces the official harness scores on 99.6\% of questions.
Scores are percentages.
Visual cost counts Qwen2.5-VL tokens of the opened images (28-pixel patches) relative to opening all retrieved images.
Paired bootstrap intervals over questions~\cite{efron1993bootstrap} for all comparisons are in \thesupp{sec:supp:ci}.

\paragraph{Policies at budget $k$.}
\emph{Retrieval order} opens the top-$k$ retrieved images, and \emph{random} opens $k$ at random.
The \emph{general decision model} is the Valen checkpoint from which \sone is initialized, given the same input and asked which images to open, which isolates the effect of our supervision.
Two self-selection baselines ask the answering model to choose up to $k$ memories in an extra call, from the captions alone or from the captions plus the same thumbnails \sone reads. The second one is a training-free selector with exactly \sone's input.
\emph{Down-sizing} opens all retrieved images, each resized so that their total equals the tokens \ours spends at budget $k$ for that question.

\subsection{\ours lies on the accuracy--cost frontier}
\label{sec:exp:frontier}

\begin{figure*}[t]
\centering
\includegraphics[width=\linewidth]{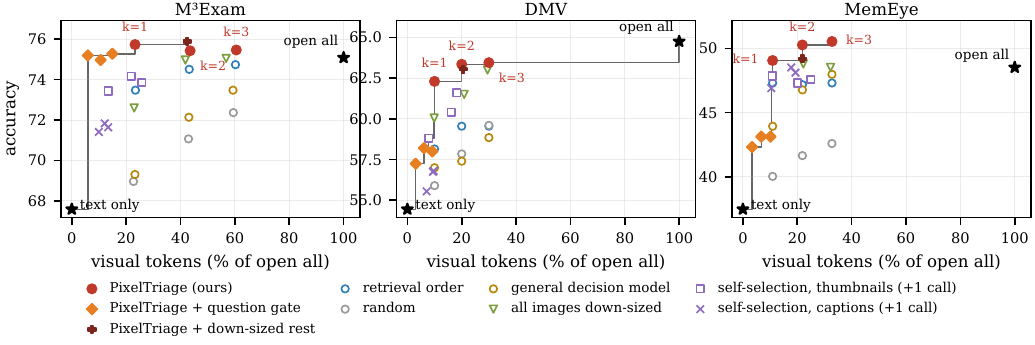}
\caption{Accuracy versus visual tokens for the evaluated delivery policies (7B answering model). Black line: accuracy--cost frontier. Self-selection makes one extra call, not counted here.}
\label{fig:pareto}
\end{figure*}

\begin{table*}[t]
\centering
\small
\setlength{\tabcolsep}{4.5pt}
\caption{Accuracy at a budget of $k$ opened images (7B answering model, 10 retrieved memories). $^\dagger$Extra answering call. Bold: best budgeted policy. \colorbox{lightgray}{Shaded}: at least as accurate as opening all.}
\label{tab:main}
\begin{tabular}{@{}l ccc ccc ccc@{}}
\toprule
 & \multicolumn{3}{c}{\mthree} & \multicolumn{3}{c}{\dmv} & \multicolumn{3}{c}{\memeye} \\
\cmidrule(lr){2-4}\cmidrule(lr){5-7}\cmidrule(l){8-10}
Delivery policy & $k{=}1$ & $k{=}2$ & $k{=}3$ & $k{=}1$ & $k{=}2$ & $k{=}3$ & $k{=}1$ & $k{=}2$ & $k{=}3$ \\
\midrule
Text proxies only & \multicolumn{3}{c}{67.58} & \multicolumn{3}{c}{54.45} & \multicolumn{3}{c}{37.47} \\
Open all retrieved images & \multicolumn{3}{c}{75.08} & \multicolumn{3}{c}{64.75} & \multicolumn{3}{c}{48.52} \\
\midrule
Random & 68.96 & 71.06 & 72.36 & 55.90 & 57.85 & 59.60 & 40.03 & 41.64 & 42.59 \\
Retrieval order & 73.47 & 74.50 & 74.73 & 58.15 & 59.55 & 59.55 & 47.30 & 47.17 & 47.30 \\
General decision model & 69.30 & 72.13 & 73.47 & 57.00 & 57.40 & 58.85 & 43.94 & 46.77 & 47.98 \\
Self-selection: captions$^\dagger$ & 71.41 & 71.83 & 71.64 & 55.55 & 56.80 & 56.75 & 46.90 & \cellcolor{lightgray}48.52 & 48.11 \\
Self-selection: + thumbnails$^\dagger$ & 73.43 & 74.16 & 73.85 & 58.80 & 60.40 & 61.60 & 47.84 & 47.30 & 47.57 \\
Down-size all images & 72.59 & 74.96 & 75.04 & 60.05 & 61.50 & 63.00 & 47.84 & \cellcolor{lightgray}48.79 & \cellcolor{lightgray}48.52 \\
\midrule
\textbf{PixelTriage (ours)} & \cellcolor{lightgray}\textbf{75.73} & \cellcolor{lightgray}\textbf{75.42} & \cellcolor{lightgray}\textbf{75.46} & \textbf{62.30} & \textbf{63.35} & \textbf{63.45} & \cellcolor{lightgray}\textbf{49.06} & \cellcolor{lightgray}\textbf{50.27} & \cellcolor{lightgray}\textbf{50.54} \\
\quad visual tokens vs.\ open all & 23\% & 44\% & 61\% & 10\% & 20\% & 30\% & 11\% & 22\% & 33\% \\
\bottomrule
\end{tabular}
\end{table*}

\cref{fig:pareto} places all 27 evaluated policies by mean visual cost and accuracy.
On every benchmark the frontier contains \ours configurations, and all 27 configurations of retrieval order, random selection and the general decision model are dominated.
On \mthree, the single image chosen by \ours reaches 75.73 at 23\% of the visual tokens, above opening all images (75.08), and on \memeye it reaches 49.06 at 11\%, against 48.52, so on both it dominates opening all images.
On \dmv, opening all images remains the most accurate policy (64.75). \ours with two images reaches 63.35 at 20\% of the tokens, 98\% of that accuracy, and the difference is not significant, whereas one image is 2.45 points lower.
The other frontier points are variants of \ours, self-selection with its extra call, and down-sizing on \dmv, which is 0.1\% cheaper than \ours and 2.3 points less accurate.

\cref{tab:main} compares the policies at equal budgets.
\ours is the most accurate budgeted policy in all nine columns.
Its margin over retrieval order is largest on \dmv, 4.2 points at $k{=}1$, where similar products make the top retrieved image a poor guess.
Self-selection by the answering model trails even when it sees the same thumbnails as \sone, by 3.5 points on \dmv at $k{=}1$, although it calls a model three times larger.

\paragraph{Choosing images versus down-sizing all of them.}
Down-sizing keeps every retrieved image visible at the same token budget.
\ours outperforms it in all nine budget settings, most clearly at the smallest budget, by 3.1 points on \mthree and 2.3 on \dmv at $k{=}1$, where down-sizing loses 2.5 and 4.7 points to opening all images.
As the budget grows, each down-sized image keeps more detail and the gap narrows, to less than half a point on \mthree and \dmv at $k{=}3$.
A question usually asks about a detail in one image, and spreading the tokens over all images blurs that detail.
Opening \ours's top image at full resolution and the others down-sized performs on par with \ours at about twice the cost.

\subsection{Transfer across memory systems and answering models}
\label{sec:exp:transfer}

\begin{table}[t]
\centering
\small
\setlength{\tabcolsep}{4.2pt}
\caption{\textbf{Transfer} to other memory systems (7B answering model) and to a 397B answering model. NaiveRAG stores text only. Bold: better of Rank@1 and Ours@1.}
\label{tab:transfer}
\begin{tabular}{@{}llcccc@{}}
\toprule
Setting & Benchmark & Text & Rank@1 & Ours@1 & All \\
\midrule
UniversalRAG & \dmv & 19.45 & 20.60 & \textbf{25.05} & 25.55 \\
UniversalRAG & \memeye & 32.61 & 39.08 & \textbf{42.72} & 41.91 \\
NaiveRAG & \dmv & 55.05 & 62.90 & \textbf{66.65} & 68.10 \\
NaiveRAG & \memeye & 31.94 & 36.25 & \textbf{40.16} & 41.51 \\
\midrule
Qwen3.5-397B & \mthree & 67.2 & 77.1 & \textbf{78.6} & 80.0 \\
Qwen3.5-397B & \dmv & 62.2 & 71.5 & \textbf{77.9} & 84.6 \\
\bottomrule
\end{tabular}
\end{table}

\paragraph{Memory systems.}
We plug \ours into UniversalRAG~\cite{yeo2025universalrag}, which routes retrieval across modalities, and into NaiveRAG, a text-only memory system to which the plug-in adds pixels (\cref{sec:plugin:hosts,tab:transfer}).
With both systems, \ours's single image outperforms the top retrieved image by 3.6 to 4.5 points on \dmv and \memeye.
For the text-only system, one image chosen by \ours raises \dmv accuracy from 55.05 to 66.65, close to adding all ten images (68.10), so the plug-in also gives a text memory system access to visual evidence at a tenth of the visual cost.
On \mthree neither system passes the 5-point criterion (gains of 2.5 and 1.8 points), and there \ours's single image stays within 0.3 points of opening all images.

\paragraph{A larger answering model.}
With Qwen3.5-397B answering and \sone unchanged, opening images matters more: all images add 12.8 points on \mthree and 22.4 on \dmv.
\ours's image outperforms the top retrieved image by 1.6 points on \mthree and 6.5 on \dmv, and a random image by 6.2 and 11.8 points (\cref{fig:recovery}), so labels from a 27B model transfer to a stronger answering model.
One image, however, no longer matches opening all images (1.4 and 6.7 points lower). Budgets above one image were not evaluated with this model.

\subsection{Cost}
\label{sec:exp:cost}

\begin{table}[t]
\centering
\small
\setlength{\tabcolsep}{4pt}
\caption{Online cost per question at $k{=}1$, including \sone, over 60 questions per benchmark without prefix caching. The time ratio is open all over ours, and compute is prefill FLOPs of ours over open all, extrapolated for 397B.}
\label{tab:cost}
\begin{tabular}{@{}llccc@{}}
\toprule
 & Answering & Open all & Ours@1 & Ratio \\
\midrule
\multirow{2}{*}{\dmv time (s)} & 7B & 2.05 & 0.71 & 2.9$\times$ \\
 & 27B & 2.77 & 1.07 & 2.6$\times$ \\
\multirow{2}{*}{\mthree time (s)} & 7B & 0.99 & 0.90 & 1.1$\times$ \\
 & 27B & 2.62 & 2.41 & 1.1$\times$ \\
\midrule
\multirow{3}{*}{\dmv TFLOP} & 7B & 238 & 65 & 27\% \\
 & 27B & 670 & 179 & 27\% \\
 & 397B-A17B & 422 & 117 & 28\% \\
\multirow{3}{*}{\mthree TFLOP} & 7B & 97 & 58 & 60\% \\
 & 27B & 296 & 162 & 55\% \\
 & 397B-A17B & 186 & 106 & 57\% \\
\bottomrule
\end{tabular}
\end{table}

\sone is a 2.25B-parameter model that runs one forward pass of about 10\,TFLOP per question. It adds 0.32 to 0.34\,s in the measurement of \cref{tab:cost} and 0.17 to 0.23\,s after warm-up.
Including this overhead, answering a \dmv question with one opened image is 2.9 times faster than opening all ten with the 7B answering model and 2.6 times faster with the 27B model, and total prefill compute falls to 27--28\% (\cref{tab:cost}).
On \mthree, where only 4.2 images are retrieved on average, compute falls to 55--60\% but the latency difference is not significant.
Retrieval order needs no decision model and answers faster than \ours (0.39 against 0.71\,s on \dmv with the 7B model) at 4.15 points lower accuracy.
Writing the note of an image takes 2.4 to 3.4\,s once, at storage time (\supp{sec:supp:write}).

\section{Analysis}
\label{sec:analysis}

\subsection{Pixel gains concentrate in one or two memories}
\label{sec:exp:sparse}

\begin{figure}[t]
\centering
\includegraphics[width=\linewidth]{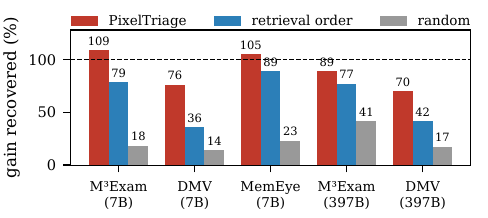}
\caption{Share of the all-image gain recovered by opening one image.}
\label{fig:recovery}
\end{figure}

\cref{fig:recovery} measures how much of the gain of opening every retrieved image one opened image recovers.
With the 7B answering model, the image chosen by \ours recovers more than the full gain on \mthree and \memeye and 76\% on \dmv, while a random image recovers at most 23\%.
Because these shares use our own selector, they bound from below how much of the gain one image can carry, and the gap to retrieval order (36\% on \dmv) shows that which image is opened decides most of the gain.
\dmv is the hardest case for retrieval order: its retrieved memories are similar products whose text never names the cue, so retrieval similarity says little about which photo shows it.
The concentration in the development labels (\cref{sec:intro}) partly reflects their construction with one referenced image per query (\cref{sec:method:data}).

\subsection{The ability comes from counterfactual supervision}
\label{sec:exp:b1}

The general decision model has the same backbone, decision head and input as \sone and was trained on general decision data only.
On the synthetic development set its question-level AUROC is 0.46, close to chance, and opening its top image yields an expected answering accuracy of 52.7, against 78.8 for \sone.
On the benchmarks it trails \ours by 5.1 to 6.4 points at $k{=}1$ and falls below retrieval order (\cref{tab:main}).
The choice of which memory deserves pixels comes from the counterfactual labels.
The auxiliary relevance output helps this learning: removing it lowers the development top-1 hit rate of $\hat g$ from 88.4\% to 80.3\% (\supp{sec:supp:ri}).

\subsection{What System One needs to see}
\label{sec:exp:input}

\begin{table}[t]
\centering
\small
\setlength{\tabcolsep}{3.5pt}
\caption{Input ablation on the development set (mean $\pm$ s.d. over three training seeds). Ref.\ helps: AUROC of $\hat g$ for whether opening the referenced image helps.}
\label{tab:input}
\begin{tabular}{@{}lccc@{}}
\toprule
\sone input & $\pvis$ AUROC & Top-1 is ref. (\%) & Ref.\ helps \\
\midrule
Memory headers & 0.718$\pm$.003 & 41.1$\pm$0.4 & 0.666$\pm$.005 \\
+ text notes & 0.809$\pm$.005 & 75.4$\pm$0.8 & 0.782$\pm$.002 \\
+ thumbnails & \textbf{0.835}$\pm$.001 & \textbf{86.7}$\pm$3.4 & \textbf{0.847}$\pm$.005 \\
\bottomrule
\end{tabular}
\end{table}

\cref{tab:input} trains \sone with three inputs, three seeds each.
With only the memory headers (session date and dialogue text), the memory with the highest $\hat g$ is the referenced one in 41\% of development queries.
Adding the text notes raises this to 75\%, and adding the thumbnails to 87\%.
The thumbnails also help beyond identifying the memory. Among referenced images, the AUROC of $\hat g$ for whether opening the image helps rises from 0.78 with notes to 0.85 with thumbnails.
This ablation was run on an earlier version of the episodes, in which questions refer to memories by date only.

\subsection{Deciding whether to look}
\label{sec:exp:gate}

The question-level probability $\pvis$ decides whether any image is opened.
We use it as a question-level budget: \ours's top image is opened only for the fraction of questions with the highest $\pvis$, and the rest are answered from text proxies.
The threshold is a quantile of $\pvis$ and needs no answers.
Estimating it on a disjoint half of the questions, or on only 50 unlabeled questions, changes the average accuracy by at most 0.4 points compared with estimating it on all test questions (\supp{sec:supp:gate}).
On \mthree, where text proxies already answer most questions, opening images for 30\% of the questions reaches 75.12 with 6\% of the visual tokens, the accuracy of opening all images. Choosing the same share of questions at random gives 70.03, and opening for 50\% reaches 75.69 with 11\% of the tokens.
On \dmv and \memeye most questions need pixels and a question-level gate has little to skip. At a 30\% opening rate it is 0.35 and 1.05 points above random gating and 7.6 and 6.5 points below opening all images.
A probability threshold fixed on the synthetic development set does not transfer across benchmarks (it opens 25\%, 77\% and 83\% of the questions), so the gate is set by its opening rate.

\subsection{When \ours does not help}
\label{sec:exp:boundary}

\paragraph{When text suffices.}
On Mem-Gallery the benchmark captions already carry the needed detail. Opening all images adds 1.7 points, and \ours's single image is 1.05 points below the top retrieved image.
SMMBench shows at most 3.7 points of headroom.
A delivery decision has little to gain in such settings, and comparing opening all images with opening none on a sample of questions identifies such settings.

\paragraph{When one image is not enough.}
Matching the accuracy of opening all images also requires informative proxies for the unopened memories. On \memeye with our short notes or no captions, one opened image is 4.45 and 4.85 points below opening all images, although still above the top retrieved image.
On \dmv, replacing an opened image's caption by its pixels deletes text that often names the cue, and one opened image then scores 53.1, below text only (54.4), with retrieval order dropping in the same way (\supp{sec:supp:conditions}).
Among \dmv questions that text alone gets wrong, \ours's image is fully correct where the top retrieved image is not in 39 cases, and the reverse holds in 10. A typical loss follows a word in a proxy, such as a wind chime with ``blue chimes'' opened for a \emph{blue orchid stem}.

\section{Conclusion}
\label{sec:conclusion}

A small model that never reads a full-resolution image can decide, before the answering model runs, which retrieved memories deserve pixels.
Because the benefit of pixels concentrates in one or two memories, opening the right ones keeps the accuracy of opening all images at a fraction of the visual tokens, and a frozen answering model supplies the counterfactual supervision needed to find them.
The decisions transfer across memory systems and to a larger answering model, for which one image no longer matches opening all images.
The same kind of decision could be made for other modalities that have a cheap proxy and an expensive raw form, such as audio and video.

{
    \small
    \bibliographystyle{plainnat}
    \bibliography{main}
}

\clearpage
\appendix
\renewcommand{\thetable}{A\arabic{table}}
\renewcommand{\thefigure}{A\arabic{figure}}
\setcounter{table}{0}
\setcounter{figure}{0}
\section{Significance tests}
\label{sec:supp:ci}

\cref{tab:ci} lists paired bootstrap intervals over questions for every comparison in \cref{tab:main} and for the variants on the frontier of \cref{fig:pareto}.
\cref{tab:ci_transfer} gives the intervals for the transfer experiments of \cref{tab:transfer}.
With the 397B model, Ours@1 outperforms Random@1 by 6.2 [$+$4.1, $+$8.3] points on \mthree and 11.8 [$+$9.6, $+$14.0] on \dmv.

\begin{table*}[t]
\centering
\footnotesize
\caption{Ours@$k$ minus each policy, in accuracy points, with 95\% paired bootstrap intervals (10{,}000 resamples). Bold: interval excludes zero.}
\label{tab:ci}
\begin{tabular}{@{}llccc@{}}
\toprule
Ours@$k$ minus & $k$ & \mthree & \dmv & \memeye \\
\midrule
Open all retrieved images & 1 & $+$0.65 [$-$0.34, $+$1.68] & \textbf{$-$2.45 [$-$4.55, $-$0.30]} & $+$0.54 [$-$3.10, $+$4.18] \\
 & 2 & $+$0.34 [$-$0.54, $+$1.26] & $-$1.40 [$-$3.25, $+$0.40] & $+$1.75 [$-$1.48, $+$5.12] \\
 & 3 & $+$0.38 [$-$0.27, $+$1.11] & $-$1.30 [$-$3.10, $+$0.45] & $+$2.02 [$-$1.08, $+$5.12] \\
\midrule
Retrieval order & 1 & \textbf{$+$2.26 [$+$1.07, $+$3.52]} & \textbf{$+$4.15 [$+$2.45, $+$5.90]} & $+$1.75 [$-$1.21, $+$4.72] \\
 & 2 & $+$0.92 [$-$0.19, $+$2.06] & \textbf{$+$3.80 [$+$2.20, $+$5.45]} & $+$3.10 [$+$0.00, $+$6.33] \\
 & 3 & $+$0.73 [$-$0.19, $+$1.64] & \textbf{$+$3.90 [$+$2.30, $+$5.55]} & \textbf{$+$3.23 [$+$0.40, $+$6.06]} \\
\midrule
Random & 1 & \textbf{$+$6.77 [$+$4.89, $+$8.75]} & \textbf{$+$6.40 [$+$4.35, $+$8.50]} & \textbf{$+$9.03 [$+$5.39, $+$12.80]} \\
 & 2 & \textbf{$+$4.36 [$+$2.71, $+$6.04]} & \textbf{$+$5.50 [$+$3.45, $+$7.60]} & \textbf{$+$8.63 [$+$4.85, $+$12.40]} \\
 & 3 & \textbf{$+$3.10 [$+$1.80, $+$4.51]} & \textbf{$+$3.85 [$+$2.00, $+$5.80]} & \textbf{$+$7.95 [$+$4.18, $+$11.86]} \\
\midrule
General decision model & 1 & \textbf{$+$6.42 [$+$4.66, $+$8.33]} & \textbf{$+$5.30 [$+$3.30, $+$7.35]} & \textbf{$+$5.12 [$+$1.21, $+$9.03]} \\
 & 2 & \textbf{$+$3.29 [$+$1.83, $+$4.89]} & \textbf{$+$5.95 [$+$3.95, $+$8.00]} & $+$3.50 [$+$0.00, $+$7.14] \\
 & 3 & \textbf{$+$1.99 [$+$0.65, $+$3.36]} & \textbf{$+$4.60 [$+$2.75, $+$6.50]} & $+$2.56 [$-$0.67, $+$5.80] \\
\midrule
Self-selection from captions & 1 & \textbf{$+$4.32 [$+$2.75, $+$5.96]} & \textbf{$+$6.75 [$+$4.95, $+$8.60]} & $+$2.16 [$-$1.08, $+$5.39] \\
 & 2 & \textbf{$+$3.59 [$+$1.95, $+$5.35]} & \textbf{$+$6.55 [$+$4.55, $+$8.65]} & $+$1.75 [$-$1.48, $+$4.99] \\
 & 3 & \textbf{$+$3.82 [$+$2.22, $+$5.58]} & \textbf{$+$6.70 [$+$4.65, $+$8.80]} & $+$2.43 [$-$0.67, $+$5.53] \\
\midrule
Self-selection from captions + thumbnails & 1 & \textbf{$+$2.29 [$+$1.15, $+$3.52]} & \textbf{$+$3.50 [$+$1.75, $+$5.30]} & $+$1.21 [$-$1.75, $+$4.18] \\
 & 2 & $+$1.26 [$-$0.04, $+$2.64] & \textbf{$+$2.95 [$+$1.40, $+$4.55]} & \textbf{$+$2.96 [$+$0.13, $+$5.93]} \\
 & 3 & \textbf{$+$1.61 [$+$0.27, $+$3.02]} & \textbf{$+$1.85 [$+$0.35, $+$3.40]} & $+$2.96 [$+$0.00, $+$6.06] \\
\midrule
All images down-sized & 1 & \textbf{$+$3.13 [$+$1.64, $+$4.63]} & \textbf{$+$2.25 [$+$0.25, $+$4.30]} & $+$1.21 [$-$2.29, $+$4.72] \\
 & 2 & $+$0.46 [$-$0.57, $+$1.49] & \textbf{$+$1.85 [$+$0.05, $+$3.70]} & $+$1.48 [$-$1.89, $+$4.85] \\
 & 3 & $+$0.42 [$-$0.61, $+$1.49] & $+$0.45 [$-$1.15, $+$2.10] & $+$2.02 [$-$1.08, $+$5.26] \\
\midrule
Its top-1 + others down-sized & 1 & $-$0.15 [$-$1.15, $+$0.88] & $-$0.75 [$-$2.60, $+$1.10] & $-$0.13 [$-$3.50, $+$3.23] \\
\midrule
With question gate (30\%) & 1 & $+$0.54 [$-$0.19, $+$1.30] & \textbf{$+$5.05 [$+$3.35, $+$6.85]} & \textbf{$+$6.74 [$+$3.23, $+$10.24]} \\
 & 2 & $+$0.46 [$-$0.38, $+$1.30] & \textbf{$+$5.15 [$+$3.35, $+$7.05]} & \textbf{$+$7.14 [$+$3.23, $+$11.05]} \\
 & 3 & $+$0.19 [$-$0.65, $+$1.07] & \textbf{$+$5.45 [$+$3.60, $+$7.35]} & \textbf{$+$7.41 [$+$3.50, $+$11.32]} \\
\bottomrule
\end{tabular}
\end{table*}

\begin{table}[h]
\centering
\footnotesize
\setlength{\tabcolsep}{3pt}
\caption{Paired differences in the transfer experiments, with 95\% bootstrap intervals.}
\label{tab:ci_transfer}
\begin{tabular}{@{}llcc@{}}
\toprule
Setting & Benchmark & Ours@1 $-$ Rank@1 & Ours@1 $-$ All \\
\midrule
UniversalRAG & \dmv & $+$4.45 [$+$3.00, $+$5.95] & $-$0.50 \\
UniversalRAG & \memeye & $+$3.64 [$+$0.40, $+$6.87] & $+$0.81 \\
NaiveRAG & \dmv & $+$3.75 [$+$2.05, $+$5.55] & $-$1.45 \\
NaiveRAG & \memeye & $+$3.91 [$+$0.94, $+$7.01] & $-$1.35 \\
Qwen3.5-397B & \mthree & $+$1.6 [$+$0.2, $+$3.0] & $-$1.4 [$-$2.5, $-$0.3] \\
Qwen3.5-397B & \dmv & $+$6.5 [$+$4.6, $+$8.3] & $-$6.7 [$-$8.2, $-$5.3] \\
\bottomrule
\end{tabular}
\end{table}

\section{Question-level gating}
\label{sec:supp:gate}

\paragraph{Setting the opening rate without the test questions.}
The gated points in \cref{fig:pareto} open \ours's top images for the 30\% of questions with the highest $\pvis$, the quantile being taken over all test questions without their answers.
\cref{tab:gate_budget} estimates the quantile instead on questions that are not evaluated: on one random half, applied to the other half and vice versa (cross-fit), or on 50 random questions, applied to the rest.
Accuracies are simulated from the recorded answers (an opened question takes its \topk{1} answer, a closed one its text-only answer), which reproduces the recorded 30\% gate within 0.1 to 0.5 points. Each estimate is repeated over 1,000 random draws.
The held-out estimates match the transductive ones, and the gate clearly outperforms random questions only on \mthree.

\begin{table}[h]
\centering
\footnotesize
\setlength{\tabcolsep}{2.5pt}
\caption{Opening rate set without the evaluated questions ($k{=}1$): mean accuracy and 2.5--97.5 percentiles over 1,000 draws. Open all: 75.08 / 64.75 / 48.52.}
\label{tab:gate_budget}
\resizebox{\linewidth}{!}{%
\begin{tabular}{@{}llccccc@{}}
\toprule
 & Rate & All test & Cross-fit & 50 unlabeled & Random & Cost \\
\midrule
\multirow{2}{*}{\mthree} & 30\% & 75.08 & 75.12 [75.00, 75.23] & 75.17 [74.38, 76.03] & 70.03 & 6.0\% \\
 & 50\% & 75.65 & 75.69 [75.65, 75.80] & 75.70 [75.04, 76.37] & 71.66 & 10.8\% \\
\multirow{2}{*}{\dmv} & 30\% & 57.15 & 57.15 [56.95, 57.40] & 57.25 [55.84, 58.53] & 56.80 & 3.0\% \\
 & 50\% & 58.65 & 58.72 [58.50, 58.95] & 58.71 [57.84, 59.63] & 58.38 & 5.0\% \\
\multirow{2}{*}{\memeye} & 30\% & 41.78 & 41.99 [41.37, 43.00] & 42.18 [38.78, 45.64] & 40.94 & 3.4\% \\
 & 50\% & 45.69 & 45.54 [45.01, 46.09] & 45.64 [43.30, 47.98] & 43.26 & 5.6\% \\
\bottomrule
\end{tabular}}
\end{table}

\paragraph{A fixed probability threshold.}

\cref{tab:gate} reports the gate with the threshold fixed on the development set ($\pvis\ge0.209$), opening \ours's top image for the questions above the threshold.
The random gate opens the top retrieved image for the same number of randomly chosen questions.
The same threshold opens images for very different shares of the questions on the three benchmarks, which is why the main text reports gating by opening rate.

\begin{table}[h]
\centering
\footnotesize
\setlength{\tabcolsep}{3pt}
\caption{Gating with a threshold fixed on the development set.}
\label{tab:gate}
\resizebox{\linewidth}{!}{%
\begin{tabular}{@{}lcccccc@{}}
\toprule
 & Opened & Cost & Gate & Random gate & All & Gate $-$ random \\
\midrule
\mthree & 25\% & 4.9\% & 75.00 & 69.63 & 75.08 & $+$5.37 [$+$3.90, $+$6.91] \\
\dmv & 77\% & 7.7\% & 61.40 & 60.48 & 64.75 & $+$0.92 [$+$0.08, $+$1.73] \\
\memeye & 83\% & 9.4\% & 47.57 & 47.12 & 48.52 & $+$0.45 [$-$1.05, $+$1.85] \\
\bottomrule
\end{tabular}}
\end{table}

\section{Relevance as an auxiliary output}
\label{sec:supp:ri}

We trained two models from the same initialization, data and seed, with and without the relevance output $\hat r_i$, on 32 GPUs with the same global batch.
With relevance, the memory with the highest $\hat g_i$ is the referenced one in 88.38\% of development queries, against 80.28\% without it ($+$8.10 [$+$6.72, $+$9.47]). The gap is 10.65 points in-source and 4.32 points cross-source.
The AUROC of $\pvis$ is unchanged (0.822 against 0.817, $+$0.005 [$-$0.002, $+$0.012]).
As a plug-in on \mthree at $k{=}1$, the model with relevance scores 75.84 and the model without it 75.31 ($+$0.54 [$-$0.11, $+$1.26]).
Removing the relevance question also changes the averaging of the per-question losses from three terms to two, so this comparison removes a task rather than isolating its loss weight.

\section{Write-time cost}
\label{sec:supp:write}

The plug-in caches, for every stored image, a thumbnail and a note written by Qwen3.5-27B.
We re-ran note writing for all 1,231 distinct images of \dmv and \mthree, one request at a time with prefix caching disabled: 2.42\,s and 91 output tokens per image on \dmv (967 images) and 3.39\,s and 134 output tokens on \mthree (264 images).
This cost is paid once per image, when it is stored, and online answering does not repeat it.
A memory system that already stores notes of this kind adds only the thumbnails.

\section{Additional conditions}
\label{sec:supp:conditions}

\paragraph{Down-sizing without captions on \mthree.}
On \mthree an opened image replaces its caption, so down-sizing replaces every caption by a small image while \ours keeps the captions of unopened memories.
With all captions removed for both policies, \ours outperforms down-sizing by 3.13 [$+$1.72, $+$4.59] points at $k{=}1$, 0.80 [$-$0.27, $+$1.87] at $k{=}2$ and 0.80 [$-$0.23, $+$1.87] at $k{=}3$, the same pattern as in \cref{tab:main} (3.13, 0.46 and 0.42).

\paragraph{Caption and delivery conditions.}
\cref{tab:conditions} varies what the unopened memories carry (the benchmark captions, our 27B short notes, or nothing) and whether an opened image keeps its caption (kept) or replaces it (replaced).
The main results use the first row of each benchmark.
\ours's image outperforms the top retrieved image in ten of the eleven conditions. The exception is \dmv with replaced captions ($-$0.6 [$-$2.6, $+$1.4]).
Whether one image matches opening all images depends on what the unopened memories keep.
On \dmv without any captions, one image outperforms all ten (64.15 against 59.85, $+$4.3 [$+$1.5, $+$7.3]).
On \memeye with our short notes or no captions, one image is 4.45 and 4.85 points below opening all images, and three images close most of the gap.
On \dmv with replaced captions, the benchmark captions, which are not filtered for the cue, often name it. Replacing the opened image's caption deletes that text, and both retrieval order and \ours fall below text-only answering at $k{=}1$.

\begin{table}[h]
\centering
\footnotesize
\setlength{\tabcolsep}{2.6pt}
\caption{Caption and delivery conditions (7B answering model). Kept or replaced: whether an opened image keeps its caption.}
\label{tab:conditions}
\resizebox{\linewidth}{!}{%
\begin{tabular}{@{}llcccccc@{}}
\toprule
 & Captions & Text & All & Rank@1 & Ours@1 & Ours@2 & Ours@3 \\
\midrule
\multirow{3}{*}{\mthree} & benchmark, replaced & 67.58 & 75.08 & 73.47 & 75.73 & 75.42 & 75.46 \\
 & our notes, replaced & 67.97 & 75.04 & 73.43 & 75.42 & 75.31 & 75.11 \\
 & none & 67.01 & 75.00 & 73.39 & 75.69 & 75.84 & 75.80 \\
\midrule
\multirow{4}{*}{\dmv} & benchmark, kept & 54.45 & 64.75 & 58.15 & 62.30 & 63.35 & 63.45 \\
 & benchmark, replaced & 54.40 & 60.00 & 53.70 & 53.10 & 56.35 & 57.20 \\
 & our notes, replaced & 28.45 & 60.00 & 49.30 & 60.15 & 61.70 & 60.35 \\
 & none & 1.35 & 59.85 & 49.80 & 64.15 & 63.35 & 59.40 \\
\midrule
\multirow{4}{*}{\memeye} & benchmark, kept & 37.47 & 48.52 & 47.30 & 49.06 & 50.27 & 50.54 \\
 & benchmark, replaced & 37.87 & 49.19 & 45.42 & 47.44 & 47.84 & 49.87 \\
 & our notes, replaced & 26.42 & 49.06 & 40.30 & 44.61 & 47.17 & 49.19 \\
 & none & 23.72 & 49.33 & 41.24 & 44.47 & 47.30 & 48.38 \\
\bottomrule
\end{tabular}}
\end{table}

\section{Protocol details}
\label{sec:supp:protocol}

\paragraph{Benchmarks.}
\mthree is converted to the Mem-Gallery dialogue format. A round that shares several images or a PDF (rendered page by page) keeps its text with the first image and adds one image-only round per further image.
The file-name questions are removed because the memory view never shows file names, leaving 654 open-ended questions.
\dmv-Bench probes are replayed from its generator: each chain has five sessions, a session browses about twelve products of one category in 22 to 28 steps, and every view becomes one memory round with the product name, price and storefront text, which do not mention the cue, and the product photo, which shows it.
The benchmark's own captions are kept as text proxies.
Answers are product names.
\memeye uses its open-ended form with absolute image paths.

\paragraph{Judges.}
The Mem-Gallery rubric is the harness prompt with JSON output, rounded to 0, 0.5 or 1, and our concurrent implementation agrees with the official harness on 99.6\% of questions.
\mthree uses the five-level rubric with Qwen2.5-VL-32B, the judge model of its release.

\paragraph{Self-selection baselines.}
The answering model receives the retrieved memories with their captions (and, for the thumbnail variant, the same 336-pixel thumbnails \sone reads) and is asked for the identifiers of at most $k$ memories whose images it needs. It may name fewer.
At $k{=}1$ it opens 0.42, 0.71 and 0.91 images on average on \mthree, \dmv and \memeye from captions, and 0.56, 0.78 and 0.94 with thumbnails.

\paragraph{Down-sizing.}
For each question the budget is the number of visual tokens of the images \ours opens at $k$.
Every retrieved image is resized, preserving its aspect ratio, to a multiple of the 28-pixel patch so that the images share the budget equally, and an image already below its share is kept.
The mean token count of the down-sized deliveries is within 6.2\% of \ours's in every setting.

\section{Implementation details}
\label{sec:supp:impl}

\paragraph{System One.}
Qwen3.5-2B with the Valen decision head (bilinear projections of dimension 256), initialized from a Valen checkpoint fine-tuned on Valen-General-100k.
Training updates LoRA adapters (rank 32, $\alpha{=}64$, learning rate $5\times10^{-5}$) on all linear layers of the language model, the vision merger ($10^{-5}$) and the decision head ($2\times10^{-4}$) for two epochs, with 28 states per optimizer step on four A100 GPUs and a maximum state length of 16,384 tokens.
The hybrid linear-attention layers of Qwen3.5 do not support custom attention masks, so the decision questions are concatenated after the memories in a fixed order, identically in training and inference.

\paragraph{Labels.}
The answering model's turn is prefilled with ``Answer:'' so that the next token is an option letter. Without the prefill, contexts with images often begin with a sentence, which depresses the letter probabilities unevenly across contexts.
Training labels use one option order, and development labels average two orders.

\section{Limitations}
\label{sec:supp:limits}

Training questions are synthetic multiple-choice recall questions built from single-image VQA, and the labels come from one answering model.
Answers are scored by language-model judges, \dmv-Bench is evaluated as a static replay of its interactive protocol, and \mthree covers the two released personas.
The experiments cover text and images.

\end{document}